\documentclass[acmtog]{acmart}

\AtBeginDocument{%
  \providecommand\BibTeX{{%
    \normalfont B\kern-0.5em{\scshape i\kern-0.25em b}\kern-0.8em\TeX}}}

\newcommand{\cmt}[1]{}
\long\def\ignorethis#1{}

\newcommand{\vc}[1]{\ensuremath{\mathbf{#1}}}

\newcommand{\pctab}{\hspace{0.2in}}

\usepackage{subcaption}
\usepackage{algorithm}
\usepackage{algpseudocode}
\DeclareMathOperator*{\argmin}{arg\,min}

\usepackage{tabularx}
\usepackage{multirow}
\usepackage{array}
\usepackage{graphicx}

\begin{document}

\title{ACE: Adversarial Correspondence Embedding for Cross Morphology Motion Retargeting from Human to Nonhuman Characters}

\author{Tianyu Li}
\email{tli471@gatech.edu}
\affiliation{%
  \institution{Georgia Tech}
  \city{Atlanta}
  \country{USA}
}

\author{Jungdam Won}
\affiliation{%
  \institution{Seoul National University}
  \country{Korea}
}
\author{Alexander Clegg}
\affiliation{%
  \institution{Meta AI}
  \city{Menlo Park}
  \country{USA}
}

\author{Jeonghwan Kim}
\email{jkim3662@gatech.edu}
\affiliation{%
  \institution{Georgia Tech}
  \city{Atlanta}
  \country{USA}
}

\author{Akshara Rai}
\affiliation{%
  \institution{Meta AI}
  \city{Menlo Park}
  \country{USA}
}

\author{Sehoon Ha}
\affiliation{%
  \institution{Georgia Tech}
  \city{Atlanta}
  \country{USA}
}


\begin{teaserfigure}
    \centering
        \includegraphics[trim={0 0 0 2cm},clip,width=0.93\textwidth]{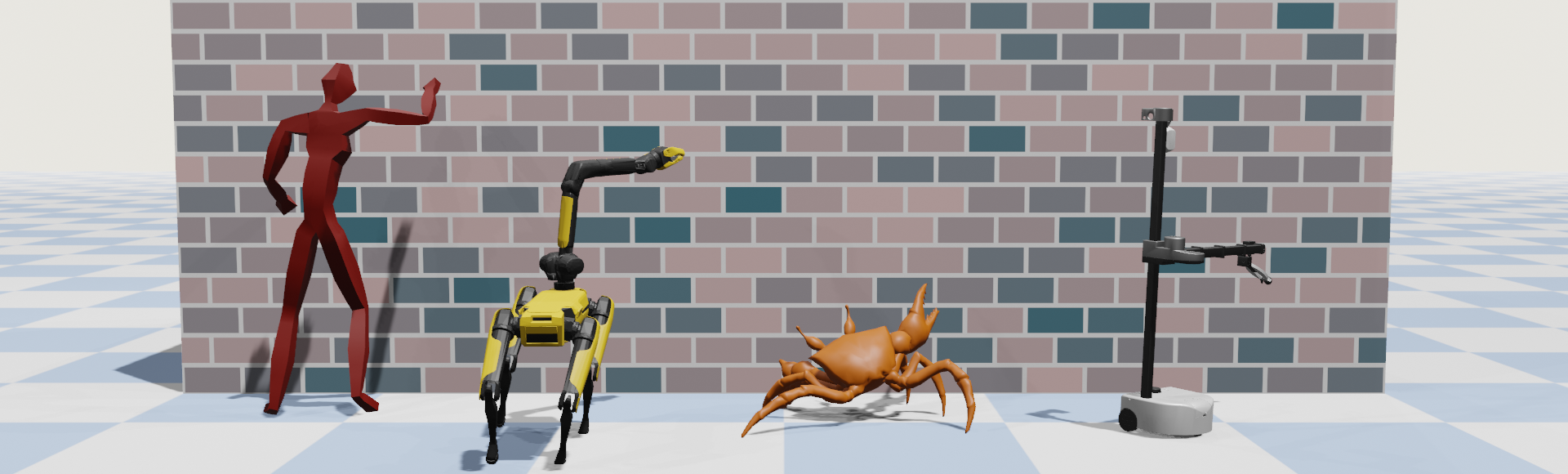}
    \caption{We propose a motion retargeting framework, Adversarial Correspondence Embedding (ACE), to retarget human motions to characters with significantly different morphologies. This figure illustrates the retargeted wall-washing motions from human (1st from left) to Spot (2nd), Crab (3rd), and Stretch (4th).}
    \label{fig:front_page}
\end{teaserfigure}

\begin{abstract}
Motion retargeting is a promising approach for generating natural and compelling animations for nonhuman characters. However, it is challenging to translate human movements into semantically equivalent motions for target characters with different morphologies due to the ambiguous nature of the problem. This work presents a novel learning-based motion retargeting framework, Adversarial Correspondence Embedding (ACE), to retarget human motions onto target characters with different body dimensions and structures. Our framework is designed to produce natural and feasible robot motions by leveraging generative-adversarial networks (GANs) while preserving high-level motion semantics by introducing an additional feature loss. In addition, we pretrain a robot motion prior that can be controlled in a latent embedding space and seek to establish a compact correspondence. We demonstrate that the proposed framework can produce retargeted motions for three different characters -- a quadrupedal robot with a manipulator, a crab character, and a wheeled manipulator. We further validate the design choices of our framework by conducting baseline comparisons and a user study. We also showcase sim-to-real transfer of the retargeted motions by transferring them to a real Spot robot.
\end{abstract}




\maketitle

\section{Introduction}

Animating non-human characters has been a longstanding topic of discussion in computer graphics. Various animation films, movies, and computer games feature beloved characters with various morphologies inspired by everyday objects (e.g., Lumière~\cite{BeautyAndTheBeast}), animals (e.g., Sebastian~\cite{LittleMermaid}), or imaginary robotic designs (e.g., Wall-E~\cite{WallE}, R2D2~\cite{Starwars}, and Omnics~\cite{Overwatch}). While moving in their own distinctive styles, these characters still need to move somewhat ``human-like'' to convey human-interpretable semantics. It is possible to use manual design, optimal control, or reinforcement learning approaches to develop motion controllers, but this process may require great manual effort from experts and take multiple iterations to produce effective and human-understandable motions, even for skilled animators. 

A more direct way to empower these non-human characters with diverse movements is to translate human motions. This motion retargeting approach not only simplifies the motion design process by avoiding complex cost or reward engineering but also has the potential to make motions human-interpretable.
However, it is not straightforward to adapt motions between characters with very different morphologies due to ambiguity and feasibility issues. For instance, which hand of a human should be mapped onto the single manipulator of a quadrupedal robot? Which types of gaits should be used when the character is following a human's walking pace? It is important to note that this is a question of style, and there is no single right answer. Even worse, some human motions are impossible for characters due to different body dimensions and structures and may cause weird dynamics or self collisions. As a result, there are fewer works that address cross-morphology motion retargeting from human to non-human characters compared to the extensive body of work on human-to-human motion mapping.

The problem of motion retargeting can be approached using a range of methods. Optimization-based motion approaches~\cite{abdul2017motion} allow us to retarget a given motion to a new character by minimizing an objective function, but they may need to be carefully tuned to take different characters or scenarios into consideration. On the other hand, data-driven approaches are able to establish implicit relationships and generalize to large scenarios. Supervised learning offers users the option to build an explicit relationship from a paired dataset~\cite{kim2022human}, but it requires precise matching of the motions between the source and target characters, which can be labor-intensive and time-consuming. On the other hand, researchers have demonstrated that recent advances in unsupervised learning can translate images~\cite{zhu2017unpaired} and language~\cite{rashid2019bilingual} across domains. Our work is motivated by these recent advances, where motion retargeting can be formulated as a translation problem between motions existing in two different domains.
While the prior works~\cite{aberman2020skeleton} investigated adversarial learning for motion retargeting between similar humanoids, it has not been extensively investigated in the context of cross-morphology motion retargeting.

In this work, we present Adversarial Correspondence Embedding (ACE), a learning-based motion retargeting framework that can translate human motions to a non-human character with significant morphological differences. The goal of our framework is to generate natural, feasible, and semantics-preserving character motions for given human motions. To this end, we build our framework on top of adversarial learning, which simultaneously trains a generator that retargets the given motion and a discriminator that evaluates the naturalness of the generated motion. We further guide the learning process by introducing an additional feature loss that preserves semantic features from the source human motion. We also pretrain motion priors that control the character's motions using a latent embedding, allowing the generator to learn a compact mapping to this latent embedding space instead of the full state of the character. 

We demonstrate that our proposed ACE framework can retarget various human motions to three very different morphologies, including a quadrupedal robot with a manipulator (Spot~\cite{bostonDynamics}), a crab character that uses two legs as manipulators, and a mobile robot with a telescopic manipulator (Stretch~\cite{Stretch}). Across such a large range of scenarios, our framework generates compelling retargeted motions that look smooth, natural, and feasible on the target character. We also compare our proposed framework against several baseline approaches through multiple objective metrics and also by conducting a user study. We further demonstrate the flexibility of our work by retargeting motions with different end-effector mappings. Finally, we showcase the sim-to-real transfer of the retargeted motions to a real Spot robot to highlight the physical validity of the proposed method.

\section{Related Works}

\subsection{Motion Retargeting}
Motion retargeting is one of the long-standing challenges in computer animation. One common approach is to formulate it as an optimization problem with different constraints on kinematic properties~\cite{gleicher1998retargetting}, end-effector motions~\cite{choi2000online} or dynamics feasibility~\cite{tak2005physically}. Although these methods can synthesize natural motions for new characters, the design of objectives and constraints often requires a labor-intensive process. 

As large mocap datasets become accessible, data-driven motion retargeting approaches have been proposed. Some researchers~\cite{delhaisse2017transfer, jang2018variational} train the retargeting function through a small set of paired data via supervised learning, then generalize to new motions. The recent success of approaches using cycle-GANs on unpaired image-to-image translation~\cite{zhu2017unpaired} inspires research on investigating motion retargeting with unpaired datasets. \citet{villegas2018neural} propose using a cycle-consistency adversarial objective with a forward kinematics-based recurrent network for motion retargeting. \citet{aberman2020skeleton} propose a skeleton-aware network to process motion such that motion can be retargeted to another character with a differently structured skeleton. However, the aforementioned data-driven methods only work for retargeting animations between humanoid characters. 

Several works~\cite{yamane2010animating, seol2013creature, rhodin2014interactive, rhodin2015generalizing, choi2020nonparametric, kim2022human} have explored retargeting motions from humans to non-humanoid characters, where the skeleton of the target character may greatly differ from the sources in terms of both structures and dimensions. These methods require selecting a few keyframe poses from human-captured motion sequences and manually pairing them with poses~\cite{yamane2010animating, rhodin2014interactive} or motion sequences~\cite{seol2013creature, rhodin2015generalizing} for the target character, which can be labor-intensive. \citet{abdul2017motion} addresses the cross-morphology motion retargeting problem by defining \emph{Groups of Body Parts} (GBPs) and translating the problem into constrained optimization to preserve the semantics of the original motion. However, the motions beside GBPs need to be designed manually. In this work, we tackle the problem of cross-morphology motion retargeting with an unpaired dataset. Our goal is to transfer the motion of a huma to a nonhuman character, while preserving the semantics. 

\subsection{Embedding Space Models for Animation and Control}
Embedding space models have been explored in controller design for both kinematic and dynamic characters. The idea is to learn a low-dimensional embedding that encapsulates natural-looking motions, then to use this learned embedding to efficiently construct controllers or further constrain output motions. Given kinematic motions, the models can be trained via supervision with specialized network architectures~\cite{zhang2018mode,starke2022deepphase,kim2022human} or in an unsupervised manner~\cite{li2021planning,ling2020character,rempe2021humor}. For physically simulated characters, the embedding models are often trained implicitly while learning imitation controllers via reinforcement learning. For instance, \citet{peng2019mcp} propose multiplicative compositional policies (MCP), which allow a policy to explore a compact embedding space to activate multiple low-level skills. \citet{won2021control} use similar embedding models to solve multi-agent problems. Recently, generative embedding models that can control simulated characters without conditioning on task-specific inputs have been studied based on conditional VAEs~\cite{won2022physics, yao2022controlvae} and adversarial learning~\cite{ASE}.
Once the embedding models are trained, a hierarchical controller can be trained to achieve various motions by traversing the embedding. 
Similar to these works, we also leverage a pre-trained motion prior, but study motion retargeting using the low-dimensional embedding space.

\section{Overview}
\label{sec:overview}

\begin{figure}
    \centering
    \includegraphics[width=0.9\linewidth]{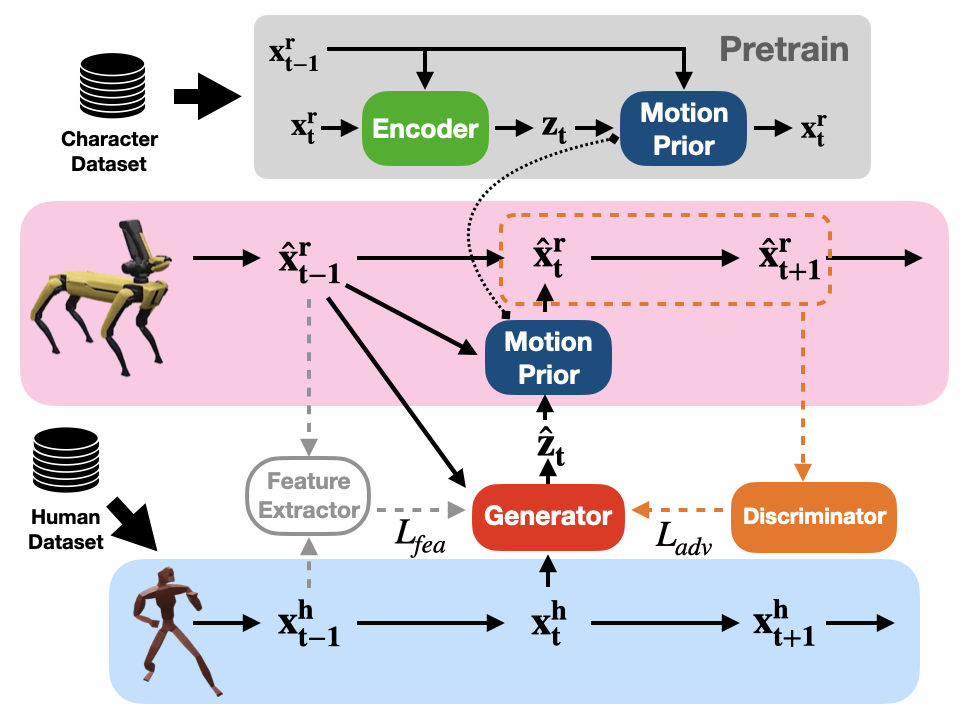}
    \vspace{-0.1cm}
    \caption{Overview diagram of Adversarial Correspondence Embedding (ACE).
    We first pre-train a motion prior that controls the character's state $\mathbf{x}^r$ with a latent variable $\mathbf{z}$. Then we train a motion retargeting Generator  that maps the current human state $\mathbf{x}^h_t$ and the previous character state $\mathbf{x}^r_{t-1}$ 
     into the latent variable, and a discriminator that determines whether the state transition is realistic or not. We also introduce an additional feature loss that guides the correspondence learning. 
    }
    \vspace{-0.6cm}
    \label{fig:overview}
\end{figure}

We present Adversarial Correspondence Embedding (ACE), a framework for retargeting human motions to characters with significantly different morphologies. Our problem takes a human motion dataset $\Omega^h$ and a character motion dataset $\Omega^r$ as inputs and learns a generator $G$ that translates a human state $\vc{x}^h$ (Section~\ref{sec:human_data}) into a target character state $\vc{x}^r$ (Section~\ref{sec:robot_data}). We aim for the generated motions to adapt to the target character's motion patterns while preserving the semantics of the original human motion. However, training such a retargeting function is not straightforward due to the significant morphological differences between the two datasets, as well as the intrinsic ambiguity of the interspecific motion retargeting problem.

We approach this problem using the framework of adversarial learning. We draw inspiration from its recent success in unpaired image-to-image translation~\cite{zhu2017unpaired} to build a correspondence between human and character motions. Unlike supervised learning~\cite{kim2022human}, unsupervised learning allows us to learn this correspondence with minimal or zero paired inputs. To further improve the quality of motion, we pre-train a motion prior $\pi$ for a character and construct a low-dimensional embedding space along with a controller that can generate diverse motions within this space. 
An overview of our approach is illustrated in Figure \ref{fig:overview}.

 \section{Pre-training of Character Motion Prior} \label{sec:motion_prior}

Motion retargeting algorithms often directly map the input motion of the source character to a high-dimensional motion of the target character. However, learning such a complex mapping can be less practical due to unstable convergence during training. Instead, we propose training a motion prior, denoted as $\pi(\mathbf{z}_t, \mathbf{x}^r_{t-1}) \mapsto \mathbf{x}^r_t$, which maps the embedded control variable $\mathbf{z}_t$ and the previous character's state $\mathbf{x}^r_{t-1}$ to the character's state at the current step $\mathbf{x}^r_t$. Later, motion retargeting is learned in this embedding space, making cross-morphology motion mapping more stable. 

Recent literature~\cite{abdul2017motion,zhang2018mode, ASE} in computer animation has discussed learning-based techniques to obtain such generative motion controllers and may complement the proposed framework. Notably, our framework is agnostic to the choice of the method used to generate the motion prior. In this paper, we use Variational Autoencoder (VAE)-based method~\cite{ling2020character} to learn motion prior $\pi$:
\begin{align} \label{equ:VAE}
    \mathbf{z}_{t} = c(\mathbf{x}^r_{t-1},\mathbf{x}^r_{t}) \\
    \argmin_{\pi}\  \mathbb{E}_{(\mathbf{x}^r_{t-1},\mathbf{x}^r_{t}) \backsim \Omega^r} ||\mathbf{x}^r_{t} - \pi(\mathbf{z}_{t}, \mathbf{x}^r_{t-1})||
\end{align}
where $\mathbf{x}^r_t$ is the character's state at time instant $t$, $\mathbf{z}_t$ is the embedded variable generated by encoding state transition $(\mathbf{x}^r_{t-1}, \mathbf{x}^r_t)$ using the encoder network $c$.  Given the embedded state $\mathbf{z}_t$ and the character's state, the motion prior $\pi$ aims to reconstruct $\mathbf{x}^r_{t}$. However, training an effective motion controller that can capture all state transitions while preserving naturalness is a challenging task. To address this, we utilize the Mode-adaptive network (MANN)~\cite{zhang2018mode} as the neual network architecture of $\pi$, which has been proven effective in previous animation works. For detailed implementation of MANN and the encoder, please refer to Sec \ref{sec:implementation}. Through joint training of the encoder $c$ and the motion prior $\pi$, we can obtain an embedded variable $\mathbf{z}$ and the motion prior for future usage.

\section{Adversarial Correspondence Embedding}
In this section, our goal is to develop an effective motion retargeting function. Although the problem of motion retargeting inherently involves ambiguity when dealing with inter-specific morphologies, we aim to achieve two notable properties in the retargeted motion. Firstly, the retargeted motion should look natural on the target character. Secondly, it should preserve the key features of the source motion. To accomplish this dual objective, we approach the problem of retargeting using generative adversarial learning~\cite{GAN,villegas2018neural,aberman2020skeleton}. This involves employing a trained discriminator $D$ to distinguish the motions retargeted by the generator $G$ from an existing character motion dataset. This encourages the generator to produce motions that are indistinguishable from pre-collected character motions, implying that they are close to the character's natural motion. Additionally, we design a simple feature loss to match the high-level features of the source and target motions. It is important to note that this feature loss plays a critical role in finding meaningful correspondence. Without it, a generator may encounter the issue of ``mode-collapse'', where it learns to produce only a limited range of motions.

\subsection{Problem Formulation}
The typical formulation~\cite{aberman2020skeleton} of motion retargeting with Generative Adversarial Networks (GANs) learns a generator $G$ to directly map the motion of the  source character $\mathbf{x}^h$ into the motion of the target character $\mathbf{x}^r$. However, such generator network can be difficult to learn due to the high dimensional state spaces of our characters. Instead, we leverage our pre-trained motion prior to learn the correspondence in an embedding space. Our generator (motion retargeting network) $G(\mathbf{x}^h_t, \mathbf{x}^r_{t-1}) \mapsto \hat{\mathbf{z}}_{t}$ takes the current human pose $\mathbf{x}^h_t$ and the previous character motion $\mathbf{x}^r_{t-1}$ to generate the embedded variable $\hat{\mathbf{z}}_{t}$. Then, the character state is produced by the pretrained motion prior $\pi(\mathbf{\hat{z}}_{t}, \mathbf{x}^r_{t-1}) \mapsto \hat{\mathbf{x}}^r_{t}$. Next a discriminator $D(\mathbf{x}^r_{t-1}, \hat{\mathbf{x}}^r_{t}) \mapsto [0, 1]$ maps the state transition to the generated dataset ($0$) or character motion dataset ($1$). 

\subsection{Training of Discriminator}
We train a discriminator $D$ to distinguish the original state transition $(\mathbf{x}^r_{t-1}, \mathbf{x}^r_t)$ in the character motion dataset $\Omega^r$ from generated transition $(\mathbf{x}^r_{t-1}, \hat{\mathbf{x}}^r_t)$ by the generator $G$, by minimizing a discriminator loss $L_D$:
\begin{align} \label{equ:discriminator_training}
    L_{D} \ = \; -\mathbb{E}_{\Omega^r}[log(D(\mathbf{x}^r_{t-1}, \mathbf{x}^r_t))] -\mathbb{E}_{\Omega^h}[log(1-D(\mathbf{x}^r_{t-1}, \hat{\mathbf{x}}^r_t))], \\
    \text{where\;}\hat{\mathbf{x}}^r_t = \pi(G(\mathbf{x}^h_t, \mathbf{x}^r_{t-1}), \mathbf{x}^r_{t-1})).
\end{align}
However, GAN uses an iterative training formulation for updating the discriminator and the generator, which often causes unstable training dynamics. One reason is non-zero gradients on the manifold of real data samples \cite{mescheder2018training} due to the approximation error in the discriminator. Thus, we incorporate a gradient penalty regularizer, as used in prior works~\cite{ASE} to stabilize the training and improve the quality of the training result. This gradient penalty augments the previous discriminator objective as:
\begin{equation}
 \label{equ: gradient penlty}
    \argmin_{D}\  \; L_D + \frac{w^{gp}}{2}\mathbb{E}_{\Omega^r}[||\nabla_\lambda D(\lambda) |_{\lambda=\mathbf{z}}   ||^2],
\end{equation}
where we set the weight term $w^{gp}$ to be $0.1$ for our experiments.

\subsection{Training of Generator} \label{sec:train_gen}
Simultaneously, we train a generator $G$ to produce natural motion transition while preserving semantic features of the source motion. It is trained by minimizing the following objective function:
\begin{equation}
  \argmin_{G}\ \; w_{adv} L_{adv} + w_{feat}L_{fea} .
\end{equation}
Here, $L_{adv}$ is the adversarial loss derived from the discriminator $D$ and encourages $G$ to generate movements that deceive $D$ into classifying them as character motion data. $L_{fea}$ is a feature loss that indicates the preservation of the semantic features from the source motion and is inspired by the concept of the group of body parts (GBP) \cite{delhaisse2017transfer}. Specifically, the adversarial loss $L_{adv}$ is calculated by:
\begin{align}
    L_{adv}(G) &=  -log(D(\mathbf{x}^r_{t-1}, \mathbf{\hat{x}}^r_{t})) \\ &=
    -log(D(\mathbf{x}^r_{t-1}, \pi(G(\mathbf{x}^h_t, \mathbf{x}^r_{t-1}),\mathbf{x}^r_{t-1} ))),
\end{align}
and $L_{feat}$ is designed to match selected high level features:
\begin{align} \label{equ: eng_loss}
    L_{fea}(G) &= || \Psi(\mathbf{x}^h_t)  - \Psi(\hat{\mathbf{x}}^r_t) || \\ &=  || \Psi(\mathbf{x}^h_t)  - \Psi(\pi(G(\mathbf{x}^h_t, \mathbf{x}^r_{t-1}), \mathbf{x}^r_{t-1})) ||,
\end{align}
where $\Psi$  is a manually designed feature function, which includes terms like end-effector positions (details in Section~\ref{sec:feature_loss}). 

To account for potential differences in the number of end-effectors between the human and target character, we have implemented a mechanism for establishing automatic correspondence between their respective end-effector indices. This is achieved by minimizing the KL-divergence between the character's end-effector position distribution and the human end-effector position distribution:
\begin{equation}
  j \mapsto i : \argmin_{i}\ KL[ \ p(x^{r,j})\ || \ p(x^{h,i})\ ]
\end{equation}
here, $j$ is the $j$-th end-effector of the target character, while $i$ is the $i$-th end-effector of the human. Besides this auto-mapping, the user can also manually define a mapping, if required.


\section{model representation} \label{sec:model}

\subsection{Human Representation} \label{sec:human_data}
We prepare a human motion dataset $\Omega^h =\{ \xi^h_1, \xi^h_2, \cdots, \xi^h_{N} \}$ that serves as an input distribution to adversarial learning. Here $\xi^h= \{\mathbf{x}^h_1, \mathbf{x}^h_2, \cdots, \mathbf{x}^h_T\}$ denotes one human motion trajectory containing states $\mathbf{x}^h$. Each trajectory can have a different length of state sequence depending on the source motion. The state $\mathbf{x}^h_t$ at time $t$ includes features as follows:
\begin{itemize}
  \item Height of the root from the ground [1 dim].
  \item Orientation of the root [4 dims].
  \item Linear and angular velocities of the root [6 dims].
  \item Position of each joint [51 dims].
  \item End-effector (foot, hand, head) position [15 dims].
   \item End-effector (foot, hand, head) velocity [15 dims].
\end{itemize}
Except for the height of the root, all the features are defined in the \emph{local coordinate} frame of the human character. Specifically, we define the local coordinate frame as follows. First, we select the pelvis of the character as the root node. Then the character's local coordinate frame is defined with its origin on the root node, its x-axis aligned with the direction that the character is facing and its z-axis is aligned with a global up-vector.

\subsection{Target Character Representation} \label{sec:robot_data}
Similar as the human motion dataset, we prepare a character motion dataset 
$\Omega^r =\{ \xi^r_1, \xi^r_2, \cdots, \xi^r_{M} \}$.
We define each motion $\xi^r = \{\mathbf{x}^r_1, \mathbf{x}^r_2, \ \cdots, \mathbf{x}^r_T\}$, where $\mathbf{x}^r$  represent a state vector. 

The character's state vector includes the following items:
\begin{itemize}
  \item Height of the root from the ground [$1$ dim].
  \item Orientation of the root [$4$ dim].
  \item Relative location of the root from the previous frame [$2$ dim].
  \item Relative orientation of the root from the previous frame [$4$ dim].
  \item Linear and angular velocities of the root [$6$ dim].
  \item Pose of each joint [joint number dim].
  \item End-effector positions [(3 * Number of EE) dim].
  \item End-effector velocities [(3 * Number of EE) dim].
\end{itemize}
Similar to Section~\ref{sec:human_data}, the features are defined in the \emph{local frame}, which is centered at the root node. The {embedded control variable} $\mathbf{z}$ of the character is defined as a latent variable with 32 dims.

\subsection{Design of Feature Loss} \label{sec:feature_loss}
\begin{figure}[t]
    \centering
    \includegraphics[width=0.45\textwidth]{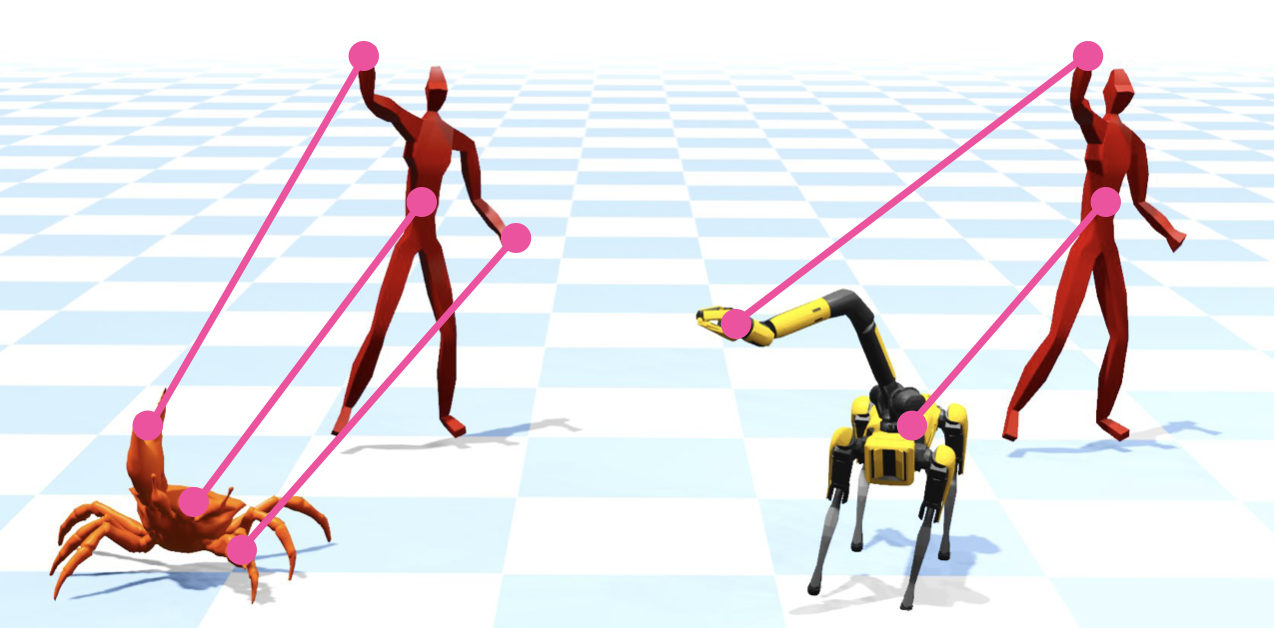}
    \vspace{-0.2cm}
    \caption{Illustration of the feature correspondence between the human and the characers. The end-effector correspondence is build in automatically. The feature loss is designed to track the root's and end-effector's motion.}
    \label{fig:feature loss}
    \vspace{-0.5cm}
\end{figure}

To preserve the semantic meaning of motions, we add an additional feature loss to the training of the generator. The features selected $\Psi$ for this feature loss are as follows:
\begin{itemize}
  \item Height of the root.
  \item Orientation of the root.
  \item Linear and angular velocities of the root.
  \item End-effector position.
\end{itemize}
All terms are normalized according to the character's body length. Figure~\ref{fig:feature loss} displays the visualization of the features. The generator is trained to match these features between the source human and target robot motion for all motions. End-effector correspondence is automatically generated using the method introduced in Section~\ref{sec:train_gen}. However, manual mapping is also possible, and we present the results in the result section. Overall, these features can be selected without requiring extensive expert knowledge and have been widely used in previous works such as \cite{aberman2020skeleton}.
 
\subsection{Network Architecture}
This works contains 4 networks: the state transition encoder $E$ and motion prior $\pi$ in the pretraining stage; the generator $G$ and the discriminator $D$ in the training stage. Here we list the their structure:
\begin{itemize}
\item Encoder Network $c$: MLP network with LeakyRelu as activation function and a structure of [512, 512, 512, 512].
\item Motion Prior $\pi$: MANN~\cite{zhang2018mode} structure with 8 experts and 512 as the unit number for each network layers.
\item Generator Network $G$: MLP network with LeakyRelu as activation function and a structure of [512, 512, 512].
\item Discriminator Network $D$: MLP network with SiLU as activation function and a structure of [512, 512, 512].

\end{itemize}

\section{experiment and evaluation}
In this section, we present qualitative evaluation of our approach on retargeting human motion to different characters. We transfer motion from a human to a Spot robot (a quadruped with a manipulator), a Crab character (a hexapod with two arms),  and a Stretch robot (a wheeled robot with a manipulator). The wide range of morphologies that we experiment with further reinforce the generality of our approach, and that it can enable generic motion transfer between inter-specific morphologies. Quantitatively, we compare our approach, ACE, against different baseline methods. In addition, we conduct a user study to further evaluate our method against other approaches. Finally, we transfer the retargeted motion to the real Spot robot.

\subsection{Implementation Details and Datasets}\label{sec:implementation}
Our motion retargeting framework is implemented in PyTorch, and the experiments are performed on a PC equipped with an NVIDIA GeForce RTX 2070 Super and AMD Ryzen 9 3900X 12-Core Processor. We optimize the parameters of the motion prior, discriminator and generator with the loss functions mentioned in Section~\ref{sec:train_gen} using the Adam optimizer~\cite{AdamOptimizer}. Training in total takes about 1 hour without any parallelization, including training the motion prior, generator and discriminator. 

To evaluate our method, we constructed a human motion dataset with 200 trajectories that contains 75594 input motion states. The human dataset is from 2 sources: CMU Motion Capture Database~\cite{CMU_Mocap}, and Ubisoft La Forge Animation Dataset ("LAFAN1")~\cite{UnisoftDataset}. The motions in the CMU dataset are around 300 frames (4.8s) while the LAFAN1 Dataset contains large motion trajectories (around 5000 frames, 90s). The selected motions include a variety of human motions including sports, dancing, housework and construction work.

The evaluation of retargeting human motions are conducted on three characters with various morphologies:
\begin{itemize}
    \item \textbf{Spot}: Spot is a quadrupedal robot~\cite{bostonDynamics} with a manipulator on its back, developed by Boston Dynamics. The robot has 32 cm long thigh and shank links. It has 12 degrees of freedom (DoFs) for locomotion and 6 DoFs for its manipulator, which results in a total of 18 DoFs.
    \item \textbf{Crab}: Crab is a hexapod character with two arms. With each leg and arm has 3 joints which leads to a 24 DoFs.
    \item \textbf{Stretch}: Stretch is a wheel-based mobile manipulation robot character from Hello Robot \cite{Stretch}. The Stretch has a single arm with four prismatic joints.
\end{itemize}

One crucial component of ACE is the learned motion prior that should be expressive and be able execute a large range of skills. This requires a diverse motion dataset on the target characters. Our character dataset contains different locomotion gaits with base linear and angular velocities ranging from -1.5 to 5m/s and -1 to 1 rad/s. The character motion are generated by rolling out kinematic controller with random target commands. For wheel-based robots, we make simplification by assuming it can walk with arbitrary speeds and directions using low-level controllers. We use a total 200k data points for each character.

\subsection{Main Results}
We illustrate various retargeted motions of all the characters in Figure~\ref{Fig:Spot Result}. Our method successfully retargeted a wide range of dynamic human motions, such as sports and construction activities. 
In all scenarios, the generated motions look feasible and preserve the original semantics of the human motions without showing significant visual artifacts, such as self-penetration or foot skating. Note that our method is general enough to support very different characters with various numbers of legs and arms.

Our system considers the different capabilities of the human and the character by matching the \emph{normalized} feature vectors. Therefore, the retargeted motions often travel or sweep less than the original human motions. For example, Spot pushes the object for $1.39$~m, which is less than  $3.48$~m of the human (Figure~\ref{fig: spot motion}, the third row). This design choice is reasonable considering the height difference between the human ($\approx1.7$~m) and the Spot robot ($\approx0.51$~m). This scaling factor can also be easily changed based on user preference. 

In our design, the characters take their footsteps in their own styles, instead of taking synchronized steps. This allows the characters to exhibit more feasible and natural motions based on their capability: for instance, the \emph{drag} motion of Spot naturally changes the gait from trotting to galloping based on the human's walking speed (\emph{1:56} in the supplemental video). However, we also want to note that this design decision may sacrifice the additional semantics of feet movements.

\subsection{Baseline Comparison}
To prove the effectiveness of the proposed method and its individual component, we compare our method to the following methods:
\begin{itemize}
\item \textbf{Neural Kinematic Network (NKN)}: the first baseline we compare is Neural Kinematic Network (NKN) of \citet{villegas2018neural}. NKN uses a recurrent neural network structure with a Forward Kinematics layer and adversarial learning with cycle consistency. However, NKN only addresses the morphology with same skeleton structure which can not be adapt to our setting as-is. Since the number of joints is different between the domains, we use the same treatment as proposed by \citet{aberman2020skeleton} which removes NKN's reconstruction loss.

\item \textbf{ACE without Feature Loss (ACEwoFea)}: the second baseline is ACE but without the feature loss. This baseline aim to evaluate how feature loss affects the result of the training.

\item \textbf{ACE without Adversarial Loss (ACEwoAdv)}: the third baseline is ACE without the adversarial loss, which corresponds to inverse kinematics only based on manual features. Here, we aim to evaluate the importance of the adversarial loss in training motion retargeting function.
\end{itemize}

To quantitatively measuring the performance of the approaches, we borrow two evaluation metrics, Diversity and Frechet Inception Distance, from the existing text-to-motion literature~\cite{guo2020action2motion,guo2022generating, tevet2022human}. We further adopt two additional metrics, a feature loss and the 
unrealistic rate.

\begin{itemize} 
\item \textbf{Diversity~(DIV)}: Diversity measures the variance of generated motion across all source human motions. From a set of all generated motions from different source human motions, two subsets of the same size $S_d$ are randomly picked. Their motion features $\{\Psi(\mathbf{x}^r_1)\cdots \Psi^(\mathbf{x}^r_{S_d})\} $ and $\{\Psi(\mathbf{x'}^r_1)\cdots \Psi(\mathbf{x'}^r_{S_d})\}$ are extracted as defined in Sec. \ref{sec:feature_loss}. The diversity of this set of motion is defined as:
$DIV = \frac{1}{S_d}\Sigma^{S_d}_{i=1} ||\Psi(\mathbf{x}^r_i) -\Psi^r(\mathbf{x'}^r_i) || .$
In motion retargeting, it is better to obtain a DIV score similar to that of the dataset. In addition, lower diversity often indicates the degeneration of the synthesized motions. 

\item \textbf{Frechet Inception Distance~(FID)}: 
FID measures the distance between feature vectors computed for two motion datasets, which is a common metric to evaluate the synthesized motion quality. We compute the feature distributions for the character's motions and the retargeted motions, and measure the  distribution difference. Lower FID indicates that two motion sets have similar feature distributions.

\item \textbf{Feature Loss ($L_{fea}$)}:We also measure the feature loss (Equation~\ref{equ: eng_loss}). Feature loss indicates the preservation of `semantic meaning'. Lower scores indicates better results.

\item \textbf{Unrealistic Frame Ratio~(UFR)}: UFR reflects the realism of the motion. It is defined as the number of generated motion frames containing unrealistic effects divided by the total number of motion frames. The unrealistic effects that we consider include self-collision, foot penetration, and foot sliding.

\end{itemize}

We evaluate our motion retargeting framework on 16 motions including sports activities, construction tasks, and house chores. The quantitative results are summarized in Table~\ref{tab:baseline result}. The qualitative results are presented in Figure \ref{Fig:Spot Result} and can be best seen in the supplementary video. 

 \vspace{-0.1cm}
\begin{table}[h]
\caption{Quantitative results on the Spot. 
}
 \vspace{-0.1cm}
\label{tab:baseline result}
\begin{tabularx}{0.45\textwidth}{ m{2cm} m{1cm} m{1cm} m{1cm} m{1cm} } 
 \hline
            & DIV $\rightarrow$ & FID $\downarrow$ & $L_{fea}$ $\downarrow$ & UFR $\downarrow$\\ [0.7ex] 
 \hline 
 Dataset & 2.254   & 0.000  & N/A  & 0.258$\%$ \\[0.7ex] 
  \hline
 \textbf{ACE(Ours)} & \textbf{2.483} & \textbf{0.489} & 0.606 & 2.071$\%$\\ [0.5ex] 
 NKN & 1.718 & 0.914 & 0.912 & 6.213$\%$\\ [0.5ex] 
 ACEwoFea & 0.445 & 0.976 & 1.975 & \textbf{0.517$\%$}\\ [0.5ex] 
 ACEwoAdv & 3.077 & 0.736 & \textbf{0.553} & 9.741$\%$\\ [0.7ex] 
 \hline
\end{tabularx}
\vspace{-0.3cm}
\end{table}

 Overall, \textbf{ACE} generates natural and compelling retargeted motions on the every chracters, including both realistic leg and arm movements (Figure~\ref{Fig:Spot Result}). The generated motions maintain the semantic knowledge in the original motion, and show minimal transfer artifacts, such as foot skating or self-collision, thanks to the pretrained motion priors and adversarial loss. The robot also shows rich whole-body movements as shown in Volleyball, Tennis and Pushing (Figure~\ref{fig: spot motion}). As a result, \textbf{ACE} outperforms all the methods in DIV (closest to DIV of the dataset) and FID, while being the second best in $L_{fea}$ and UFR.

Although \textbf{ACEwoFea} produces natural motions, the generated motions fall into the repetitive patterns and lose the semantic information of the human motion, as demonstrated by the Diversity value of 0.445 and $L_{fea}$ value of 1.975 while \textbf{ACE} gets a higher diversity and a lower feature loss. This mode collapse phenomenon is very common in adversarial learning settings. The features are used as a regularization term in training to help mitigate mode collapse.

\textbf{ACEwoAdv} is capable of generating reasonable outputs when retargeting certain human motions by preserving the semantic features: it shows the lowest $L_{fea}$ of 0.553. However, there are many scenarios where it fails to produce satisfactory results, despite the use of a pretrained motion prior. This can be demonstrated by its high FID value of 0.736 while \textbf{ACE} has 0.489. 
In addition, it often produces unrealistic motions with self-collision, foot penetration, and foot sliding, which is supported by the worst unrealistic frame ratio (UFR) value of 9.741\%.

Our comparison reveals that when generating high-quality retargeted motions, \textbf{ACE} outperforms \textbf{NKN} in many criteria. Previous work motion retargeting work~\cite{aberman2020skeleton} has noted the crucial role played by the reconstruction loss in \textbf{NKN}. Removing this component for supporting cross-morphology scenarios could result in a degradation of the motion quality produced by \textbf{NKN}.

\subsection{User Study}
In addition to the aforementioned quantitative evaluation, we conducted a user study to assess how our method performs in terms of visual perceptual quality when compared to other baseline methods. Our study group comprised 20 participants with varying levels of expertise and experience in character animation.  Prior to the study, participants were provided a comprehensive explanation of the study, without any information about the underlying method. 

In the user study, five different source human motions were randomly selected from the dataset and presented to the subjects, along with the retargeted character motions generated by each method. The selected human motions comprised various sports and other human activities. The subjects are asked to evaluate the generated motion in 0 to 5 scales based on realism and the magnitude alignment to the source human motion. During the study, the user has no access to the method that generates the motion.

The preference results are presented in Table \ref{tab:userstudy}. According to the results, most users picked the retargeted motions of \textbf{ACE} as the most favorable. On the other hand, \textbf{ACEwoFea} received the worst result as it loses all semantic information of the human motions. Our findings also show that \textbf{ACEwoAdv} achieved a relatively high score, although \textbf{ACE} is still statistically better than \textbf{ACEwoAdv} with a p-value of 0.03. This is because \textbf{ACEwoAdv} can provide better results compared to \textbf{NKN} and \textbf{ACEwoFea}. However, it is important to note that unrealistic effects such as foot sliding or foot penetration may not be easily captured by non-expert users.
The user study offers strong evidence of the effectiveness of our approach in retargeting motions across different morphologies.

\begin{table}[h]
\caption{User study on scoring the retargeted motions.}
\vspace{-0.3cm}
\begin{tabularx}{0.45\textwidth}{ m{1.6cm} m{1.8cm} m{1.8cm} m{1.7cm} } 
 \hline
    \textbf{ACE(Ours)}   & NKN & ACEwoFea & ACEwoAdv\\ [0.7ex] 
 \hline 
 \textbf{4.25} $\pm$ \textbf{0.39} & 1.41 $\pm$ 0.78 {\footnotesize	(***)} & 1.01 $\pm$ 0.80 {\footnotesize	(***)} & 3.95 $\pm$ 0.45 {\footnotesize	(*)} \\ [0.7ex] 
 \hline
\end{tabularx}
\label{tab:userstudy}
\vspace{-0.55cm}
\end{table}

\subsection{Flexibility for Incorporating User's Preference}
\begin{figure*}[t]

 \centering
    \begin{minipage}[b]{.995\textwidth}
        \centering
        \includegraphics[width=0.995\textwidth]{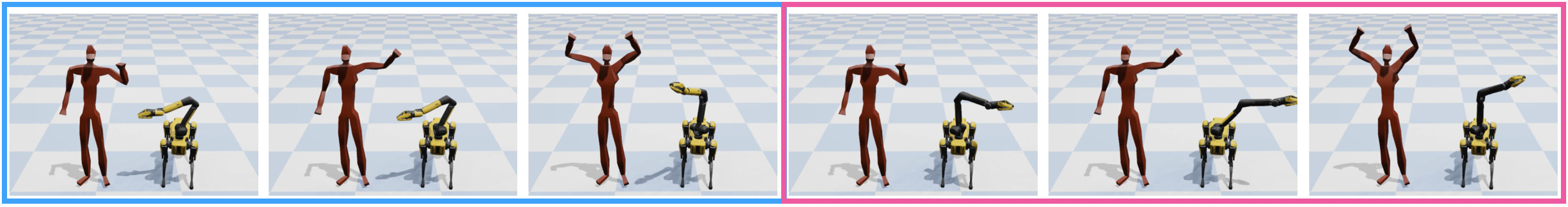}
        \vspace{-0.1cm}
    \end{minipage}
\vspace{-0.3cm}
\caption{Different generated Spot \emph{Wave} motions by varying end-effector mapping. The blue uses auto-mapping while the pink one is manually assigning Spot's manipulator with human left arm.}
\label{Fig:different mapping}
\end{figure*}

Although ACE includes an automatic end-effector mapping mechanism, the end-effector correspondence can also be manually set according to the user's preference. In the previous section, the automatic mapping assigned the manipulator of the Spot robot to the right hand of the human character. However, we can manually assign the manipulator to the left hand of the human and use ACE to produce new retargeted motions. Figure \ref{Fig:different mapping} presents the result under the new mapping. In addition to manual end-effector mapping, other heuristics, such as specific joint-level mapping or foot-pattern synchronization, can be incorporated into ACE by modifying the feature loss function.


\subsection{Real Robot Experiments}
One application of our work is to reproduce the retargeted motions on a real robot via sim-to-real transfer. This is important in robotics because it allows the robot to acquire various motor skills from human movements. Once motion is retargeted, we can use several techniques for executing the given motion, such as motion imitation~\cite{peng2018deepmimic} or model-based control~\cite{li2021fastmimic}.

We selected the Spot robot as the robotic platform. The API takes as input the base velocity command and the manipulator joint angle, which are compatible with our design of the latent command $\mathbf{z}$. Therefore, we could deploy the retargeted motion by replacing the motion prior with the vendor-provided controller. 

\begin{figure*}[t]
    
 \centering
    \begin{minipage}[b]{.995\textwidth}
        \centering
        \includegraphics[width=0.995\textwidth]{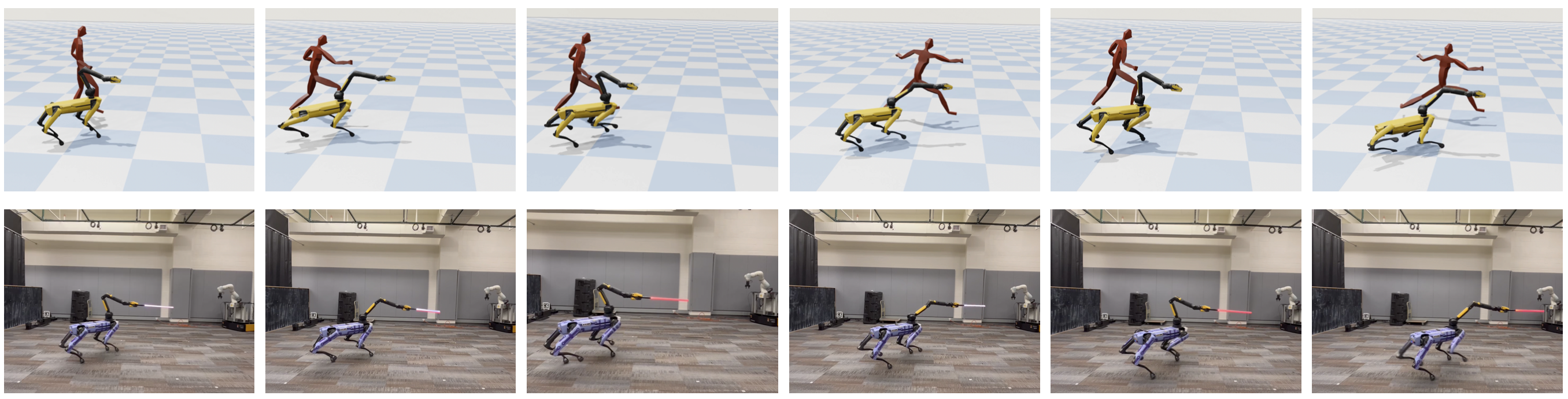}
        \vspace{-0.1cm}
        \label{Fig: HW Fencing}
    \end{minipage}

   \begin{minipage}[b]{.995\textwidth}
        \centering
        \includegraphics[width=0.995\textwidth]{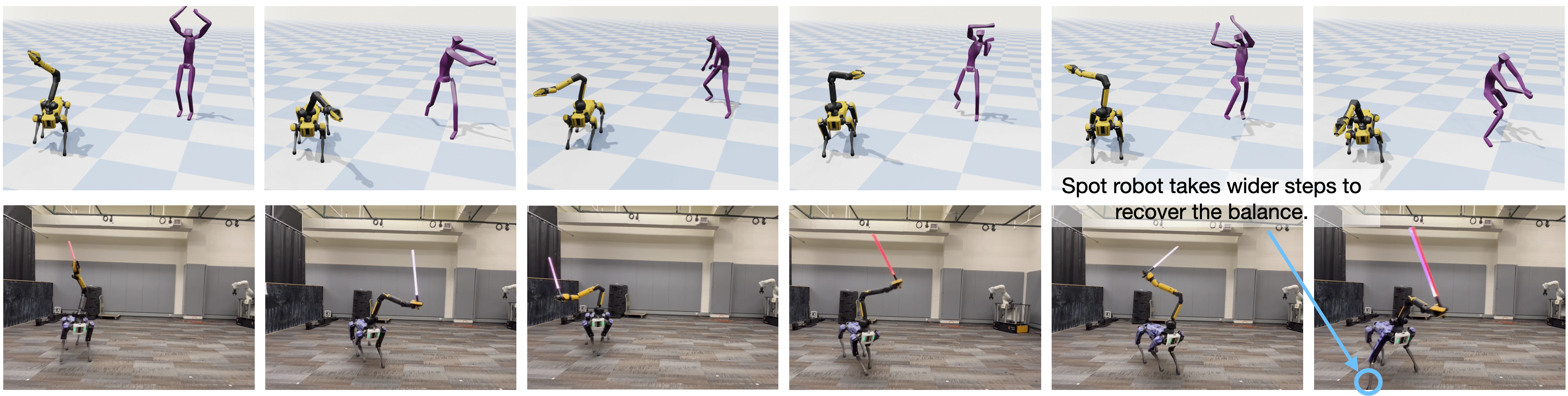}
        \vspace{-0.1cm}
        \label{Fig: HW Sword}
    \end{minipage}
    

\vspace{-0.3cm}
\caption{We successfully transferred two whole-body motions generated for Spot, \emph{Fencing} and \emph{Sword}, to the Spot robot without any failures.}
\label{fig: hardware results}
\end{figure*}

We transferred two motions, \emph{Sword} and \emph{Fencing}, which involve rich full-body motions and rapid arm motions. The Spot robot was able to execute both sequences at 100~\% success rates out of five trials. Our ACE framework generates physically plausible motions that are within the region of attraction of the given controller. However, the details of the footsteps were different, particularly when the arm is stretched and starts to affect the balance. In this case, the real Spot robot takes wider steps to recover the balance (Figure \ref{fig: hardware results}). The numbers of footsteps are also different: 14 for our kinematically retargeted motions and 26 for the real Spot due to the differences in the controllers as well as the need for balance recovery. However, it is important to note that these experiments are designed to highlight the robotic application but do not indicate any guarantee on sim-to-real transfer.

\section{Discussion and Future Work}
This work presents a learning-based framework, Adversarial Correspondence Embedding (ACE), which retargets a given human motion to another character with significant morphological differences. Our framework leverages adversarial learning to generate natural character motions while guiding the correspondence learning via a feature loss. We also introduce a pre-trained motion prior and learn retargeting in a compact embedding space, which leads to smooth, physically realistic motion on the character. We demonstrate that the proposed framework can generate compelling retargeted motions on three characters, Spot, Crab, and Stretch, with various morphologies. We also conduct baseline comparisons and a user study to justify the design decisions of our framework. Finally, we highlight the potential robotic application by transferring the retargeted motions to a real Spot robot.

Our work has a few notable limitations.
Our current implementation does not systematically consider the differences in dynamic capabilities between morphologies. For example, a larger human may be able to turn faster than a smaller character. Exploring solutions for such scenarios may require investigating long-horizon motion planning with reinforcement learning or time-warping. 
In addition, our trained model is limited to a single robot. 
We plan to investigate a general motion embedding space that can freely translate motions back and forth betweenvarious morphologies, including human to character, character to human, and character to character. 
Finally, our framework may be less effective in handling characters without clear notions of arms and legs, such as a shark~\cite{seol2013creature} or a caterpillar~\cite{rhodin2014interactive}. Extending the proposed framework for more general characters will be an interesting future research direction.
\begin{acks}
\end{acks}

\bibliographystyle{ACM-Reference-Format}
\bibliography{acm-reference}

\newpage

\begin{figure*}[t]

 \centering
    \begin{minipage}[b]{.995\textwidth}
        \centering
        \includegraphics[width=0.995\textwidth]{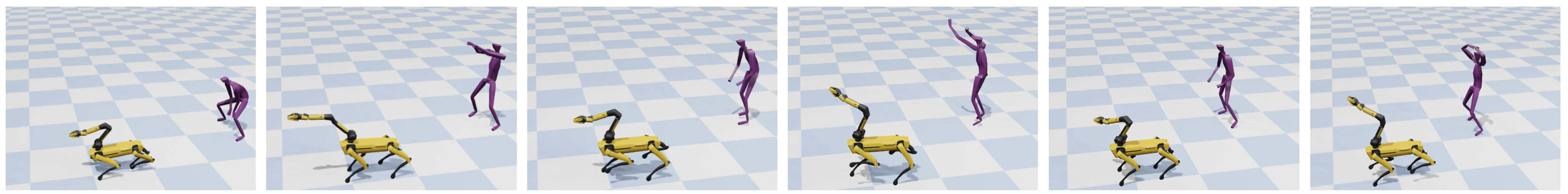}
        \vspace{-0.2cm}
    \end{minipage}

   \begin{minipage}[b]{.995\textwidth}
        \centering
        \includegraphics[width=0.995\textwidth]{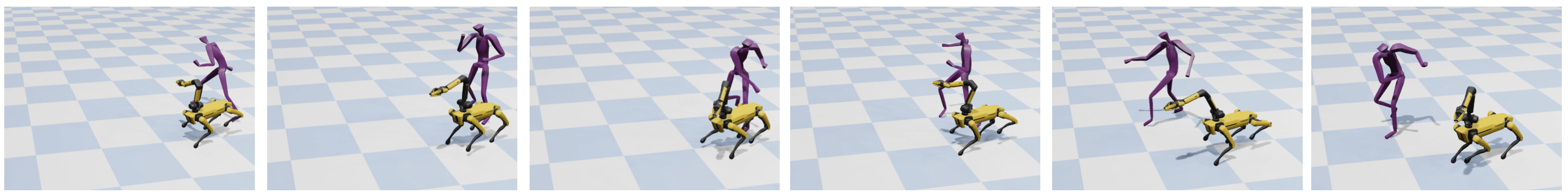}
        \vspace{-0.2cm}
    \end{minipage}
    
     \begin{minipage}[b]{.995\textwidth}
        \centering
        \includegraphics[width=0.995\textwidth]{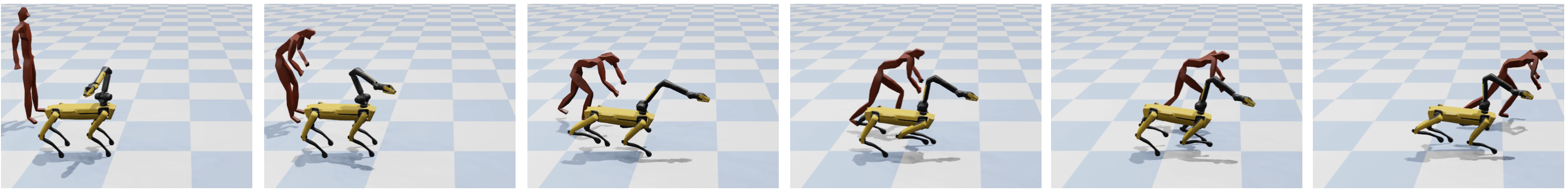}
        \vspace{-0.2cm}
        \subcaption{Spot \emph{Volleyball}, \emph{Tennis}, and \emph{Pushing}.}
        \label{fig: spot motion}
    \end{minipage}

     \begin{minipage}[b]{.995\textwidth}
        \centering
        \includegraphics[width=0.995\textwidth]{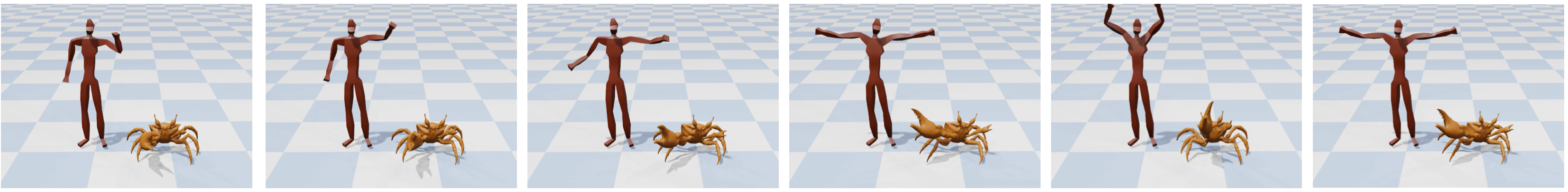}
        \vspace{-0.2cm}
    \end{minipage}

    \begin{minipage}[b]{.995\textwidth}
        \centering
        \includegraphics[width=0.995\textwidth]{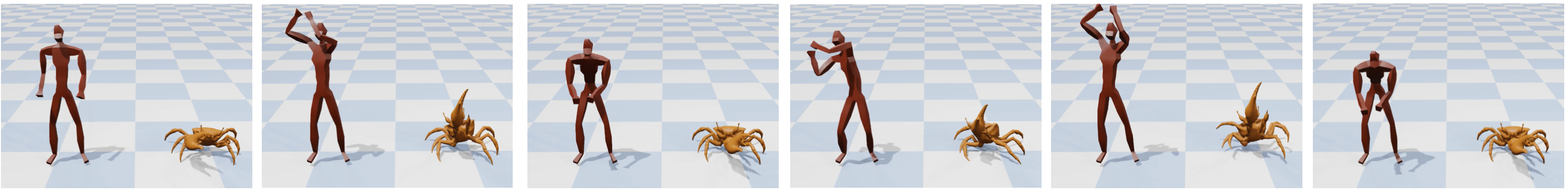}
        \vspace{-0.2cm}
        \subcaption{Crab \emph{Waving} and \emph{Chopping}.}
    \end{minipage}

     \begin{minipage}[b]{.995\textwidth}
        \centering
        \includegraphics[width=0.995\textwidth]{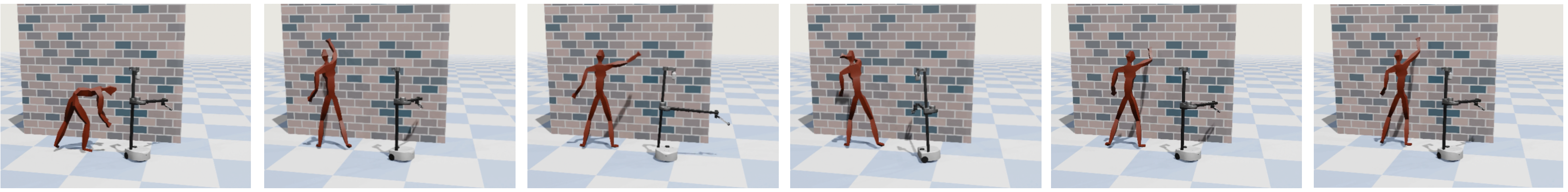}
        \vspace{-0.2cm}
    \end{minipage}

     \begin{minipage}[b]{.995\textwidth}
        \centering
        \includegraphics[width=0.995\textwidth]{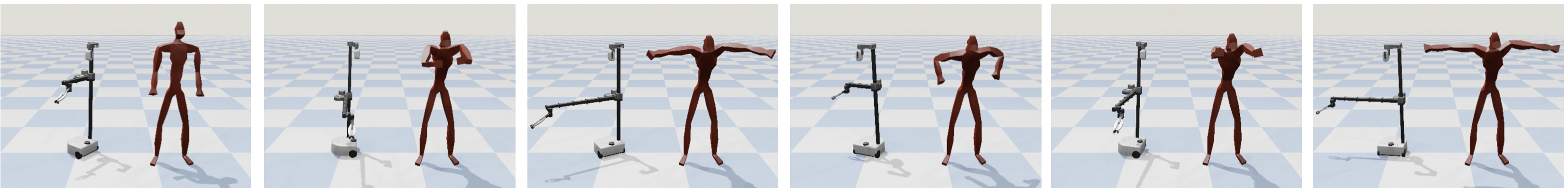}
        \vspace{-0.2cm}
        \subcaption{Stretch \emph{Wall Washing} and \emph{Swimming}.}
    \end{minipage}
    
 \vspace{-0.3cm}
\caption{Retargeted Motions on Spot, Crab and Stretch. Our framework can retarget various human motions while preserving semantic features.}
\label{Fig:Spot Result}
\end{figure*}

 \vspace{-0.5cm}

\begin{figure*}[t]

 \centering
    \begin{minipage}[b]{.995\textwidth}
        \centering
        \includegraphics[width=0.995\textwidth]{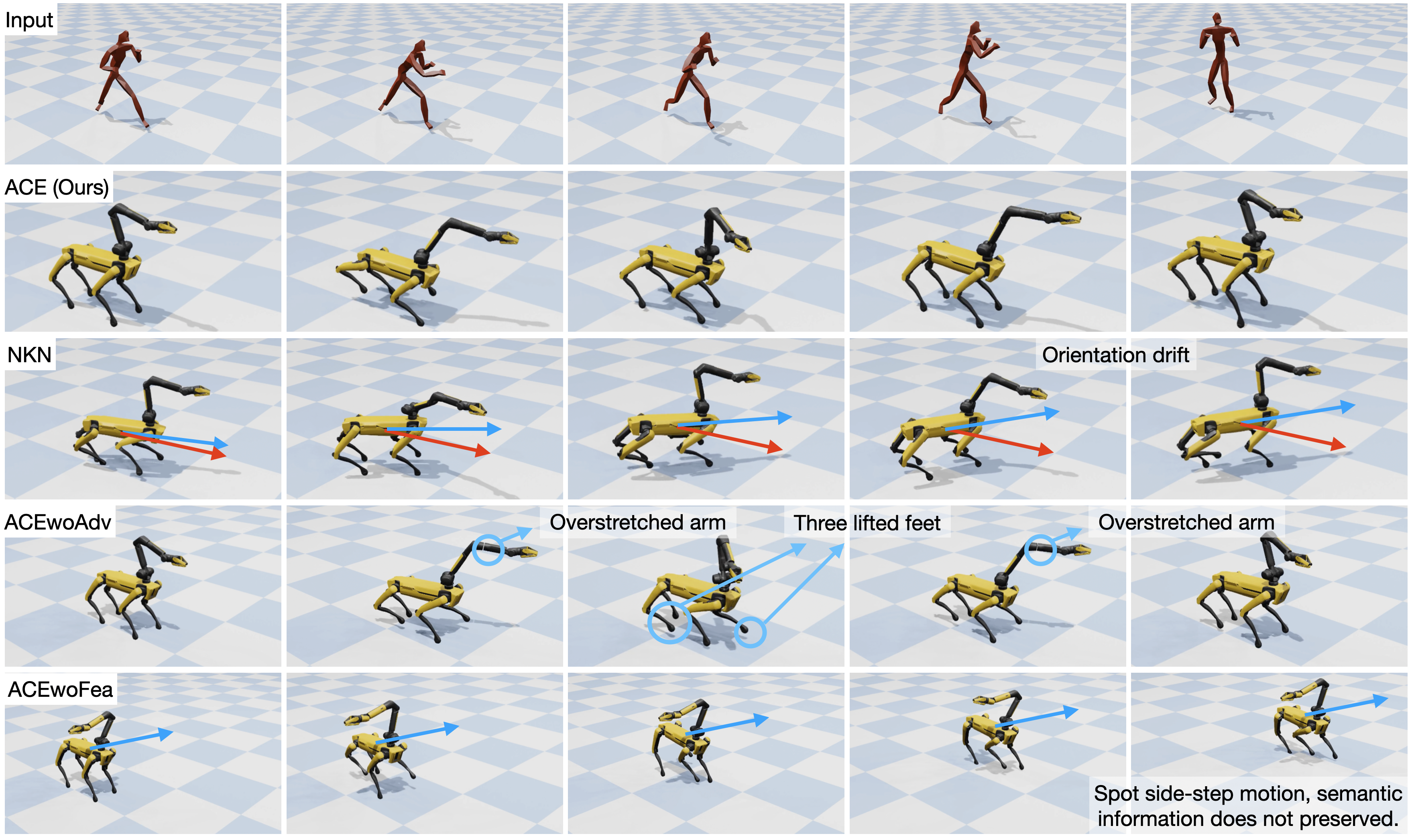}
        \vspace{-0.1cm}
    \end{minipage}

    

    
\vspace{-0.3cm}
\caption{Comparing our method with the baseline methods, we demonstrate that our approach can generate realistic character motion while preserving the semantic information of the input motion. }
\label{Fig:baseline comparison}
\end{figure*}

\end{document}